# Pass-Fail Criteria for Scenario-Based Testing of Automated Driving Systems

**Robert Myers**
*New Mobility Technologies Directorate*
*Connected Places Catapult*
Milton Keynes, UK
robert.myers@cp.catapult.org.uk

**Zeyn Saigol**
*New Mobility Technologies Directorate*
*Connected Places Catapult*
Milton Keynes, UK
zeyn.saigol@cp.catapult.org.uk

*Abstract*— The MUSICC project has created a proof-of-concept scenario database to be used as part of a type approval process for the verification of automated driving systems (ADS). This process must include a highly automated means of evaluating test results, as manual review at the scale required is impractical.

This paper sets out a framework for assessing an ADS's behavioural safety in normal operation (i.e. performance of the dynamic driving task without component failures or malicious actions). Five top-level evaluation criteria for ADS performance are identified. Implementing these requires two types of outcome scoring rule: prescriptive (measurable rules which must always be followed) and risk-based (undesirable outcomes which must not occur too often). Scoring rules are defined in a programming language and will be stored as part of the scenario description.

Risk-based rules cannot give a pass/fail decision from a single test case. Instead, a framework is defined to reach a decision for each functional scenario (set of test cases with common features). This considers statistical performance across many individual tests. Implications of this framework for hypothesis testing and scenario selection are identified.

## 1 Introduction

Regulatory verification of an automated driving system (ADS) is a complex technical problem. The process must consider many different causes of potential harm, which are commonly grouped into categories such as 'cybersecurity', 'functional safety' or 'behavioural safety'. This paper proposes an approach for assessing behavioural safety, i.e. the vehicle's performance of the dynamic driving task under normal operating conditions.

The space of possible scenarios which an ADS may have to respond to is large and complex, even in comparison to the advanced driver assistance systems discussed in ISO/PAS 21448:2019. A promising approach to ADS verification involves evaluating performance over a large set of scenarios [1]: the required scale and need for reproducible outcomes points towards extensive testing in simulation [2].

Once a vehicle has been tested, the results of the test must be evaluated. Manual review is probably too subjective to be the primary tool used, and the time required would be excessive. Automated evaluation is the only feasible option, which creates the need for a pass/fail framework based on machine readable rules. This is made challenging since existing regulations often rely on subjective comparisons (e.g. comparing behaviour to a "careful and competent" driver [3]) or impose constraints which are expected to be ignored under certain circumstances (it is societally, if not legally, acceptable to break most road rules if the alternative is a collision). For automated evaluation of results, a pass/fail framework needs to allow minor deviations from driving norms in critical situations, while maintaining a high level of performance overall.

MUSICC was funded by the UK Department for Transport. The ideas expressed in this paper are those of the authors only.

This remainder of this paper starts with a summary of the project this work was conducted as part of, MUSICC (Multi User Scenario Catalogue for CAVs) (Sec. 2), and a review of relevant prior art (Sec. 3). The novel contributions are a concise statement of the criteria by which an ADS should be evaluated (Sec. 4), and a framework for using these (Sec. 5), which applies prescriptive rules for some criteria and statistical measures for others. This framework has two parts: a method for evaluating performance in a single scenario (Sec. 6), and then, for the statistical measures, a method for evaluating the pass/fail decision for a full set of scenario tests (Sec. 7). Finally, conclusions are presented in Sec. 8.

## 2  Background: MUSICC Project

The MUSICC project has created a proof-of-concept system for use in type approval of Connected and Autonomous Vehicles (CAVs). At the heart of the system is a scenario library: a searchable database storing scenarios and metadata. It supports scenario import/export, searching by ODD, basic editing, and a scenario approvals process with version control. All features can be accessed through a web interface; searches and downloads can also be performed using an API. Fig. 1 shows the inputs and outputs to/from the system.

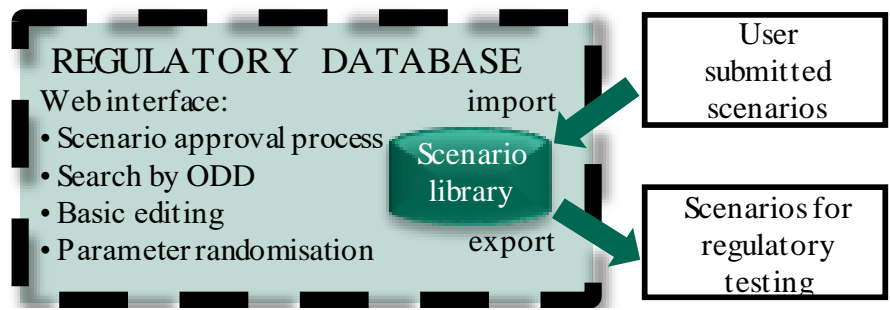

*Fig. 1:Diagram of the MUSICC system*

MUSICC categorises scenarios according to the three levels of abstraction defined by PEGASUS [4]. These are: functional (a human readable description of what happens in a scenario), logical (a fully defined scenario, but where some values may fall within a range) and concrete (a fixed value exists for every variable). All three types of scenario can be represented in the database:

- Logical scenarios are stored as a combination of an OpenSCENARIO [5] format XML file (which defines a concrete scenario) and parameter stochastics stored in a separate XML file.
- Functional scenarios are represented by using a tag to link logical scenarios derived from the same functional description.
- Concrete scenarios are the lowest level of abstraction: in MUSICC, they can be generated automatically from a logical scenario. It is also possible to store them directly if desired.

MUSICC has completed a beta phase, with users from over 50 organisations spread over 12 countries trialling the system and participating in an advisory group for the project, and is now available as Open-Source software [6]. However, to our knowledge, it has not yet been used as part of a real ADS test. A working model for pass/fail criteria is a key part of making MUSICC useful, and for regulatory use, these criteria must be fair, clear and objective. They must also be transparent, so that both ADS developers and the wider public can have confidence in the testing. This paper is intended to share our work in this area with the wider community.

## 3 Previous Work

Relevant previous work has identified some essential components of a testing framework, criteria which results should be evaluated against, and different approaches to scoring. This section summarises the key findings.

ADS developer FiveAI has proposed a general outline for CAV type approval [7]. They identify the need for a regulatory database of scenarios, contributed by industry and stored centrally (which is the core functionality of the MUSICC system). They also note a need for "a publicly described test oracle" to determine whether a test has passed or failed, and suggest it should take the form of a "Digital Highway Code" (DHC). The code would contain a detailed, measurable specification for good driving, applicable across a wide range of scenarios.

Reference [8] also takes the approach of testing scenarios in simulation and gives three criteria to be used when making pass/fail decisions as an example (maintaining safety distance, not causing collisions, and avoiding collisions caused by others). It identifies a need for a risk assessment to be made when evaluating results but does not go into detail about how this process should work.

Intel Mobileye have produced a paper setting out a concept called "Responsibility Sensitive Safety" [9]. This gives a formal, mathematical definition of basic driving principles ("reasonable behaviour"). If these principles are followed by all road users, it can be proven that no collision should result. If there is a crash, the road user(s) that behaved unreasonably is responsible.

The concept of using an appropriately weighted scoring system to measure performance is introduced by [10] in the context of training a machine learning system. A metric for performance is used which considers crashes, safety margins, traffic laws, comfort and achieving the objective specified.

References [11] and [12] identify metrics to evaluate the risk in scenarios. In [12] exposure, severity and controllability are used to select tests based on the risk to a human driver. Reference [11] uses two types of metric, some relating to near collisions by the vehicle under test (e.g. time to collision) and others relating to the impact on traffic flow. Calculated scores are used to classify scenarios by criticality; however, the paper does not discuss how to determine if performance in critical scenarios is adequate.

The more complex a system becomes, the more possible combinations of inputs and outputs there are which may be relevant to its response. Reference [13] explores methods to reduce the parameter space by performing tests of an ADS's subsystems. This includes a process for converting high level safety goals into low level requirements for each subsystem, which can be measured in testing.

Our work builds on the state-of-the-art by fully defining a comprehensive approach to measuring the behavioural safety of an ADS. It works for all scenario types and can provide evidence of compliance with top level performance requirements.

## 4 ADS Evaluation Criteria

We have identified two key points relating to ADS performance which are not covered by previous work:

- Some undesired behaviours (e.g. following too close to the car in front) may be acceptable if the alternative is a more serious outcome (e.g. a collision). This means it is necessary either to define the exact outcomes which are acceptable in each scenario or to apply a scoring system which penalises some outcomes more than others.

- The "proper response" to unreasonable behaviour is described in [9]. However, human road users go further, acting to mitigate the risk from unreasonable behaviour before it materialises. For example, drivers usually avoid stopping on bends with restricted visibility, even though other vehicles have a responsibility not to hit them if they do. This type of risk mitigation does not appear to be covered by the existing work.

Taking these together with the outcomes of the previous work, we suggest that an ADS should:

1. Never cause a collision.

2. Drive in a way which allows for the possibility of unreasonable behaviour by other road users. While it may not always be possible to avoid a collision, the risk should be mitigated (this will have to be balanced against the need to make progress).

3. Obey traffic rules. It is possible that some changes in the law will be made to accommodate CAVs, though the nature these is unknown and may differ between countries. We note that human drivers would not typically be prosecuted for breaking rules to avoid a crash, even when specific legal exemptions do not exist.

4. Leave reasonable safety margins[1] where possible (e.g. do not drive close to the limit of dynamic performance, leave large gaps when passing obstacles). Note that an ADS could trade one of these for an increased margin on another (e.g. harsh braking in order to increase gap with obstacle).

5. Behave in a way which is considerate to other road users. This means avoiding behaviour likely to disrupt traffic flow or confuse a human driver.

Some of these criteria, such as "never cause a collision", can be expressed as a mathematical or logical calculation. Prescriptive tests like this are important: in this case, if all CAVs subscribe to the same definition of causing a collision, no collisions between them should occur. However, some important criteria cannot be measured in this way. For example, "drive in a way which allows for the possibility of unreasonable behaviour by other road users" requires a compromise between making progress and risk, as in some circumstances it might be acceptable for a good ADS to crash.

One approach to testing these less precisely defined criteria is to compare the performance of an ADS with a reference algorithm. This could be based on human driver performance (as in [14]) or another standard. This creates a measurable test for every scenario: the ADS under test must never perform 'worse' than the reference algorithm. However, the approach has two key disadvantages:

- The reference algorithm would be technically difficult to produce and creates its own verification problem. In effect, the regulator is required to create a driving decision making algorithm to compare against others.
- This approach could restrict innovation. Requiring that an ADS is better than the reference algorithm in every scenario (rather than a set of scenarios taken together) could force developers to use a decision-making algorithm very similar to the reference one.

Instead, for testing higher levels of autonomy, we propose that a mixture of prescriptive and risk-based tests should be used. As mentioned above, prescriptive tests evaluate well-defined rules for each scenario. Risk-based tests supplement them where it is undesirable to define acceptable behaviour for every situation. A risk-based test compares the likelihood and severity of negative outcomes to tolerability of risk criteria defined by a regulator. Methods for implementing these tests and evaluating the results are addressed in sections 6 and 7 respectively.

---

[1] Safety margins are good practice in all areas of engineering where failure represents a significant risk. In this case they allow for minor faults or errors in modelling to be tolerated without leading to a collision. Also, since humans leave safety margins, the failure of an automated vehicle to leave them is likely to concern road users.

Risk based tests should be used where most ADS are expected to perform well most of the time, but specifying exactly which scenarios an ADS should be able to handle is impractical or undesirable. Specifically, we suggest risk-based tests should be used for measuring mitigation of unreasonable behaviour, safety margins, and inconsiderate behaviour by the ADS under test.

An implication of adopting the mixed approach is that each concrete scenario may result in a fail (if a prescriptive rule is violated), but the absence of a fail does not necessarily mean performance was acceptable. The outcome of a single risk-based test will be a severity level, which will need to be aggregated over a set of scenarios to be meaningful.

Table I summarises the prescriptive and risk-based tests proposed to evaluate the criteria outlined earlier.

TABLE I.     TESTS OF EACH TYPE

| **Criterion** | **Tests** | | |
| --- | --- | --- | --- |
| | *Type* | *Description* | *Rationale* |
| Never cause a collision | Prescriptive | Where all other road users behave reasonably, no collisions are acceptable | It is reasonable for all of these crashes to be avoided. |
| Drive in a way which allows for the possibility of unreasonable behaviour by other road users. | Risk-based (statistical) | Assign a severity score based on collision severity | By defintion, it is not always possible to avoid crashes resulting from unreasonable behaviour. However, decisions made by the ADS can affect the risk. |
| Obey traffic rules (depending on reasonable behaviour of other road users) | Both | If other road users behave reasonably, traffic rules must be obeyed. If other road users behave unreasonably, assign severity level. | This is an approximation of the current de-facto approach: rules may occasionally be broken to avoid a crash caused by another road user. |
| Leave reasonable safety margins | Risk-based (statistical) | Assign severity level (always lower severity than a collision) | Safety margin violations are preferable to collisions. Allowing occasional safety margin reductions is also consistent with current driving norms (e.g. driving slightly closer to the vehicle in front to avoid harsh braking). |
| Behave considerately to other road users | Risk-based (statistical) | Assign severity level | Apparently inconsiderate or confusing behaviour (e.g. gentle braking in response to an event which is not yet visible to humans) is undesirable but may occasionally be accepted in return for other benefits (e.g. smoother traffic flow). |

## 5   Overview of Measurement Approach

We have identified two broad approaches to tackling the problem of evaluating ADS behaviour. The first involves creating a detailed, measurable specification for good driving (a 'Digital Highway Code' [7]). This is a potentially powerful approach, but substantial development effort will be required before it can be relied on for system validation.

An alternative approach involves specifying pass/fail criteria for each scenario. Knowing the circumstances where these will be applied will make them easier to define: only things which are directly relevant to the scenario need to be considered. For the remainder of this paper, we follow this approach, in line with our agile philosophy of making a system available to the community at the earliest point possible to allow wide evaluation. However, even if the stakeholders converge on the DHC approach in the future, the knowledge presented in this paper will still be relevant.

The MUSICC solution for pass/fail criteria consists of two main components:

- Outcome scoring rules, both prescriptive and risk based, which are applied to a concrete scenario. These take the form of a script stored in the MUSICC database for each logical scenario .
- A method to reach a pass/fail decision by combining severity scores from risk-based tests on many concrete scenarios.

These are explained in Sec. 6 and Sec. 7 respectively.

## 6 Outcome Scoring Rules

### 6.1 Approach

Outcome scoring rules examine each concrete scenario tested. They assign a result: this could be a fail (if a prescriptive rule is violated) but will otherwise be a severity level. We assume outcomes are binned into one of a finite number of severity levels $s_i$ – e.g. the four levels used by ISO 26262, from S0 (cosmetic damage) to S3 (life-threatening or fatal injuries). Sec. 7 explains how these scores are used to reach a pass/fail decision for each functional scenario. Scenario results are evaluated against the prescriptive and risk-based criteria from Sec 4, as shown in Fig.2.

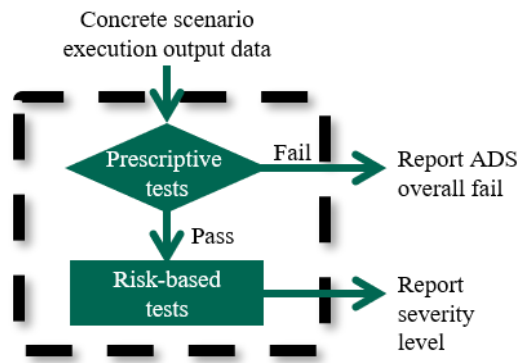

*Fig.2: Evaluation of outcome scoring rules for a single concrete scenario*

Scoring rules may need to be relatively complex, e.g.:

- A crash severity model may be required to distinguish between minor collisions and those likely to lead to serious injury.
- The correct behaviour may be different depending on the value of parameters within a logical scenario (e.g. different values for the speed of a vehicle may alter whether its behaviour is classed as reasonable). It is also desirable to be able to reuse parts of the same calculation in more than one scenario (e.g. apply the same crash severity model).

We have implemented outcome scoring rules using a programming language to accommodate this complexity in a generic manner.

For cross-simulator compatibility, a finite, standardised list of data available to the scoring rules will be needed. Items likely to be required include velocities of actors, the relative distance between any pair of actors, time-to-collision between actors. Scores may be calculated during execution or when the scenario is complete. Real time calculation could allow execution to stop early if the result of a test has been failed, but the processing time benefit of this is limited, as a failure of any prescriptive test represents a failure for the entire ADS.

### 6.2 Implementation in CARLA simulator

A proof-of-concept implementation of the outcome scoring rules has been built as an extension to the CARLA simulator [15]. It can download scenarios from the MUSICC database, apply tests, and produce a report on the results.

Data logging capability has been added to CARLA, allowing specified values to be recorded from the scenario. These are written to CSV files for evaluation by the outcome scoring rules. Defining a log file format has two key advantages over requiring the evaluation of outcome scoring rules to be built into the simulator:

- It is a relatively straightforward way of adding pass/fail criteria to existing simulators.
- If changes are made to the pass/fail criteria, results can be re-evaluated without running scenarios again (as long as no additional data needs to be logged).

Outcome scoring rules are encoded as Python scripts. The structure is designed to be compatible with existing unit test frameworks, allowing for integration into a development process. The rules are written using assertions: a library of these can be reused across any number of scenarios.

The CSV log format and Python scripts make it straightforward to reuse scoring rules with different simulators.

## 7 Reaching a Pass/Fail Decision

Fig.3 gives an overview of the process to be followed in order to reach a pass/fail decision on an ADS. A sampling strategy is used to determine which scenarios should be tested. These are tested and the outcome scoring rules discussed in Sec. 6 are applied. If these result in a prescriptive test failure, no further evaluation is required: the ADS has not met the prescribed standard. If they do not, the risk-based tests are evaluated. This remainder of this section explains how the key steps in this process work.

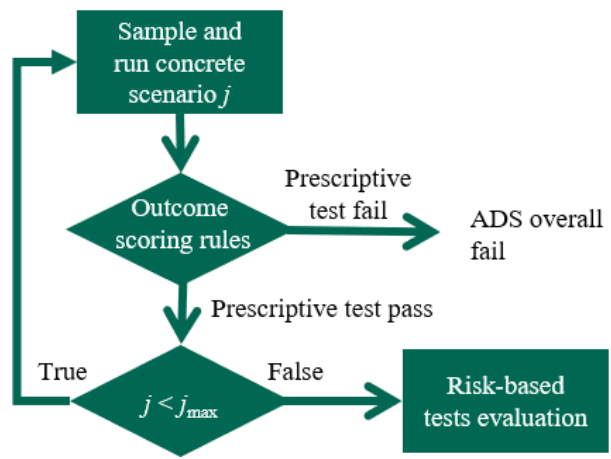

*Fig.3: Overall evaluation process*

### 7.1   Evaluation of risk-based tests

Risk assessments are appropriate where there is a need to balance cost or other aspects of performance against safety. Rather than giving prescriptive requirements for how a system should behave, they require the likelihood and severity of failures to be considered. Generically, they include:

- An assessment of how likely a hazard is to occur and the severity of harm which could result.
- A comparison of the above to tolerability of risk criteria. These are used to determine whether the level of risk posed is acceptable or not[2].

---

[2] Often these include a level where a hazard may be accepted provided that the risk is as low as reasonably practicable. The approach in this paper compares risks to tolerability criteria, but identifying mitigations is out of scope.

Examples of standards which use this type of approach include ISO 26262 (Road vehicles – Functional safety) and ISO 14971 (Risk management for medical devices). This section sets out an approach to making risk-based decisions about the safety of highly automated vehicles.

*7.1.1   Inputs*

The proposed evaluation process requires the following inputs:

- A tolerability of risk statement for each severity level, of the form: no functional scenario[3] should cause events of severity $\geq s_i$ at a rate $> \lambda_i$ per unit time that the vehicle is in use[4].
- An estimate of exposure, *e*, for each functional scenario. Exposure may be represented as a rate of occurrence (e.g. occurs once per hour for a typical vehicle) or a proportion of time driven (e.g. occurs about 1% of the time for a typical vehicle).
- A conditional probability of occurrence for each concrete scenario, *p(**x**)* (where ***x*** represents all the variables required to define a concrete scenario). Given that a functional scenario is selected, each concrete scenario occurs with probability p(***x***). The distribution of p(***x***) used in the database should approximate the real world.

We assume these inputs to be calculated externally; the small set of $\lambda_i$ values will be part of the evaluation framework, and the exposure and p(***x***) values will be stored as metadata against each scenario within MUSICC.

*7.1.2   Calculations*

Calculations in this section are explained for exposures expressed as a rate per hour, but an analogous approach can be applied where they are expressed as a proportion of time.

First, an acceptable occurrence rate $l_{i,n}^{acceptable}$ is defined for each severity level (i) in each functional scenario (n) according to (1):

$$l_{i,n}^{acceptable} = \frac{\lambda_i}{e_n} \quad (1)$$

$l_{i,n}^{acceptable}$ is dimensionless and represents a proportion of concrete scenarios.

The next step is to estimate the actual rate of occurrence of each severity level, $l_{i,n}^{actual}$. This is achieved by sampling within a functional scenario according to p(***x***), executing the resulting concrete scenarios and using the procedure from Sec. 6.1 to produce severity scores.

Finally, $l_{i,n}^{actual}$ is compared to $l_{i,n}^{acceptable}$ using the statistical tests discussed in part 7.1.3 of this section.

*7.1.3   Hypothesis testing*

The process above will result in an estimate for how often different severities of outcome will occur. However, this rate will have a confidence interval associated with it: it cannot be known with precision. This is likely to be most important for high exposure scenarios which rarely result in a high severity outcome. The acceptable rate of occurrence for these outcomes will be very low, while the difference that one test can make to the overall ADS rate estimate is large.

---

[3] A functional scenario is analogous to a hazard in traditional risk assessment terminology. The approach of applying tolerability criteria to each hazard separately is widely used in quantitative and semi-quantitative approaches.

[4] Note that this form of statement avoids situations where an ADS could avoid a failure by a higher severity score. For example, an ADS at risk of failing through violating safety margins cannot pass by having a collision instead.

Two types of hypothesis test could be applied:

- We could test the hypothesis $H_a$: "severe outcomes will occur at a rate of $l_{HighSeverity,n}^{acceptable}$ or greater", where $l_{HighSeverity,n}^{acceptable}$ is calculated as in Sec. 7.1.2. Rejecting this hypothesis would demonstrate that the vehicle is safer than the minimum requirement.
- Alternatively, we could test the hypothesis $H_b$: "severe outcomes occur at a rate of $l_{HighSeverity,n}^{acceptable}$ or less". Rejecting this hypothesis would demonstrate that the vehicle does not meet the minimum requirement.

For some values of $l_{HighSeverity,n}^{acceptable}$, even a very safe vehicle would require a lot of tests before $H_a$ can be rejected. Using $H_b$ avoids this problem, as a vehicle is assumed safe in the absence of evidence to the contrary. However, failing to prove that a vehicle is unsafe is not the same as proving that it is safe. The implications of adopting either type of test are illustrated in the example below.

Consider a functional scenario which occurs at least once on most journeys. For the purposes of this example, we assume that $l_{HighSeverity,n}^{acceptable}$ for a high severity outcome is $1 \times 10^{-7}$ per test (based on an estimate of the rate of UK fatal vehicle crashes derived from [16] and [17]: $1.7 \times 10^{-7}$ per hour driven).

Consider simulating the behaviour of this vehicle in 10 million scenarios and testing $H_a$ and $H_b$ at the P=0.05 significance level:

- In the case where there are 0 negative outcomes, testing $H_a$ gives P=0.37. This means that $H_a$ cannot be rejected: even with perfect performance, 10 million samples is not enough to demonstrate safety.
- In the case where there are 2 negative outcomes, testing $H_b$ gives P=0.18. This means that, despite the actual rate observed being substantially higher than the target, this may well occur by chance for a vehicle which meets the target level of safety.

In this example, any number of collisions between 0 and 2 means that neither $H_a$ or $H_b$ is rejected: the vehicle is neither provably safer nor less safe than the standard. This range will narrow if more common outcomes are tested.

*7.2 Sampling strategy*

In general, the more concrete scenarios which are tested, the better the estimate of overall risk which can be made. However, it is a well-documented problem that for autonomous vehicles a vast amount of testing is required to statistically demonstrate safety [2]. A key benefit of scenario-based testing is that it can be focused on the more challenging cases. For example, testing a vehicle driving on an empty road is likely to be mostly uninformative, since a mature ADS is unlikely to fail. Focusing testing on scenarios with a higher expected failure rate will give a better indication of the risk which an ADS presents. This could include a coverage driven verification approach such as that discussed in [7].

For the process outlined earlier, the selection of concrete scenarios within a functional scenario must be representative of the real world. However, the number of concrete scenarios tested per functional scenario is unconstrained. This means that functional scenarios which place a higher demand on the ADS can be prioritised. The process works best if concrete scenarios within a given functional scenario create a similar level of demand on the ADS. If this is not the case, prioritising some functional scenarios over others is pointless – all of them have a similar probability of resulting in an undesired outcome.

For the functional scenarios which result in a particular severity occurring at a low rate (e.g. the occurrence of a serious crash when driving on an empty road), a proportionately large error margin will remain. This is an inherent result of using a finite number of samples to estimate the properties of an entire test space. There are two approaches which can help to provide more confidence in these areas:

- ADS developers could adopt a doer/checker architecture, such as that discussed in [18]. This does not guarantee optimal performance, but at least ensures that the more easily verified constraints imposed by the checker are always obeyed.
- A commonly used principle in health and safety management is that the rate of minor accidents correlates with the rate of major accidents [19]. We suggest that, where the rate of serious outcomes cannot be measured, the rate of occurrence for less serious outcomes remains a useful measure of safety. In the earlier example of driving along an empty road, this might mean putting an emphasis on the number of times safety margins are eroded.

### 7.3   Prescriptive tests

Prescriptive tests are self-contained: they specify well defined rules which should never be broken. Therefore, a failure of a prescriptive test in a single scenario indicates a design flaw in the ADS. The pass/fail criteria should indicate a fail in this case, since the requirements specified have not been met. Whether regulators should accept systems with known flaws is outside the scope of this paper.

## 8   Conclusions

We have presented a framework that enables the behavioural safety of an ADS to be automatically assessed in a transparent and objective way. This is critical for regulatory approval of CAVs, given that a large number of scenario tests will be needed to have confidence in the safety of these vehicles, making human analysis of all the results unrealistic.

Our framework defines the acceptable behaviour for an ADS as a set of rules within each scenario, which makes the rulesets easy to define. We have developed a comprehensive set of criteria that ADS should be evaluated against, and proposed that some of these are treated as prescriptive, whereas others should be evaluated statistically. This imposes some additional requirements for the scenario database and regulator, but allows the scope of testing to be more comprehensive than in the alternatives we have identified. Finally, we have detailed how results should be extracted from individual scenario tests in a portable way, and how they should be aggregated over functional scenarios to arrive at a pass/fail decision for an ADS.

Choosing the right set of functional scenarios is critical for our framework to be effective. These scenarios should cover as much of the test space as possible, and be defined at a suitable level of granularity, allowing the more challenging scenarios to receive higher test coverage.

One key issue we have not addressed is exactly what the assessment rules should be – for example, when the road width and traffic situation make it safe to overtake a cyclist, or how much to slow down if the car in front is behaving erratically. Further, the required ADS safety levels need to be set by the regulators, taking into account stakeholder views (in our framework, this would correspond to setting numbers such as the acceptable risk $\lambda_i$). Opinions on how safety goals for CAVs should be defined will be sought as part of a new project (CertiCAV), which will also further examine how to apply them in practice. This could include demonstrating the framework from this paper in a proof-of-concept CAV approval process. Looking further ahead, there are many other problems to be addressed in defining a certification process for CAVs [20] [7], and we will be exploring some of these in CertiCAV and other closely-related projects.